\documentclass[final]{cvpr}

\usepackage{times}
\usepackage{epsfig}
\usepackage{graphicx}
\usepackage{amsmath}
\usepackage{amssymb}

\usepackage{multirow}
\usepackage{flushend}

\usepackage[pagebackref=true,breaklinks=true,colorlinks,bookmarks=false]{hyperref}

\def\cvprPaperID{6178} 
\def\confYear{CVPR 2021}

\begin{document}

\title{Cross-Modal Collaborative Representation Learning and a Large-Scale \\ RGBT Benchmark for Crowd Counting}

\author{First Author\\
Institution1\\
Institution1 address\\
{\tt\small firstauthor@i1.org}
\and
Second Author\\
Institution2\\
First line of institution2 address\\
{\tt\small secondauthor@i2.org}
}

\author{Lingbo Liu$^{1}$, Jiaqi Chen$^1$, Hefeng Wu$^1$, Guanbin Li$^{1,2}$, Chenglong Li $^{3}$, Liang Lin$^{1,4}$\thanks{The corresponding author is Liang Lin. Lingbo Liu and Jiaqi Chen share first-authorship.}\\
$^1$~ School of Computer Science and Engineering, Sun Yat-sen University, China\\$^2$~Pazhou Lab, Guangzhou, 510330, China{~~~}$^3$~Anhui University{~~~}$^4$~DarkMatter AI Research
}

\maketitle
\pagestyle{empty}
\thispagestyle{empty}

\begin{abstract}\label{intro}
  Crowd counting is a fundamental yet challenging task, which desires rich information to generate pixel-wise crowd density maps. However, most previous methods only used the limited information of RGB images and cannot well discover potential pedestrians in unconstrained scenarios. In this work, we find that incorporating optical and thermal information can greatly help to recognize pedestrians. To promote future researches in this field, we introduce a large-scale RGBT Crowd Counting (RGBT-CC) benchmark, which contains 2,030 pairs of RGB-thermal images with 138,389 annotated people.
  Furthermore, to facilitate the multimodal crowd counting, we propose a cross-modal collaborative representation learning framework, which consists of multiple modality-specific branches, a modality-shared branch, and an Information Aggregation-Distribution Module (IADM) to capture the complementary information of different modalities fully. Specifically, our IADM incorporates two collaborative information transfers to dynamically enhance the modality-shared and modality-specific representations with a dual information propagation mechanism.
  Extensive experiments conducted on the RGBT-CC benchmark demonstrate the effectiveness of our framework for RGBT crowd counting. Moreover, the proposed approach is universal for multimodal crowd counting and is also capable to achieve superior performance on the ShanghaiTechRGBD \cite{lian2019density} dataset. Finally, our source code and benchmark are released at {\color{blue}\url{http://lingboliu.com/RGBT_Crowd_Counting.html}}.
\end{abstract}

\section{Introduction}

Crowd counting~\cite{kang2018beyond,gao2020cnn} is a fundamental computer vision task that aims to automatically estimate the number of people in unconstrained scenes. Over the past decade, this task has attracted a lot of research interests due to its huge application potentials (e.g., traffic management \cite{zhang2017understanding,liu2020dynamic} and video surveillance \cite{xiong2017spatiotemporal}). During the recent COVID-19 pandemic \cite{velavan2020covid}, crowd counting has also been employed widely for social distancing monitoring \cite{ghodgaonkar2020analyzing}.

\begin{figure*}
\centering
   \includegraphics[width=1.945\columnwidth]{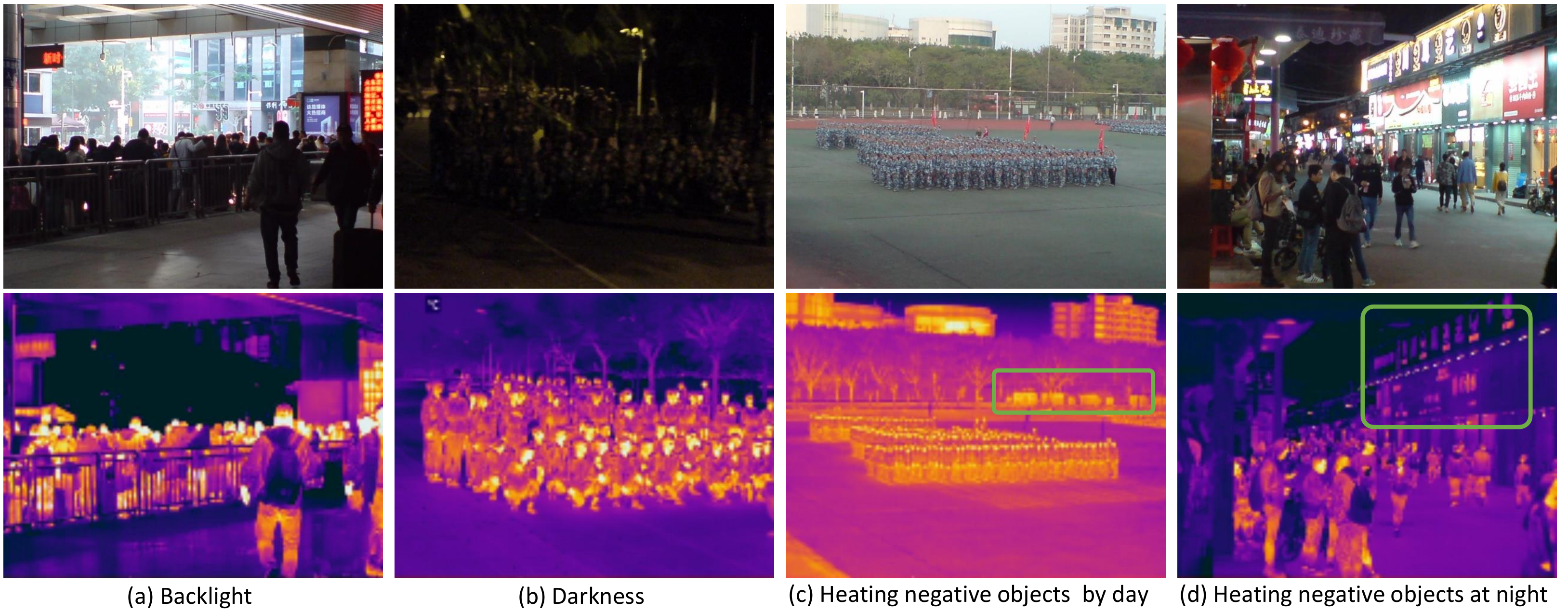}
\vspace{0mm}
   \caption{Visualization of RGB-thermal images in our RGBT-CC benchmark. When only using optical information of RGB images, we cannot effectively recognize pedestrians in poor illumination conditions, as shown in (a) and (b). When only utilizing thermal images, some heating negative objects are hard to be distinguished, as shown in (c) and (d).
   }
\label{fig:RGBT_sample}
\vspace{-2mm}
\end{figure*}

In the literature, numerous models \cite{zhang2016single,sindagi2017generating,liu2018crowd,zhang2019attentional,bai2020adaptive,
li2018csrnet,liu2019crowd,ma2019bayesian,liu2019context,liu2020semi} have been proposed for crowd counting. Despite substantial progress, it remains a very challenging problem that desires rich information to generate pixel-wise crowd density maps. However, most previous methods only utilized the optical information extracted from RGB images and may fail to accurately recognize the semantic objects in unconstraint scenarios. For instance, as shown in Fig.~\ref{fig:RGBT_sample}-(a,b), pedestrians are almost invisible in poor illumination conditions (such as backlight and night) and they are hard to be directly detected from RGB images.
Moreover, some human-shaped objects (e.g., tiny pillars and blurry traffic lights) have similar appearances to pedestrians \cite{zhang2016faster} and they are easily mistaken for people when relying solely on optical features. In general, RGB images cannot guarantee the high-quality density maps, and more comprehensive information should be explored for crowd counting.

Fortunately, we observe that thermal images can greatly facilitate distinguishing the potential pedestrians from cluttered backgrounds. Recently, thermal cameras have been extensively popularized due to the COVID-19 pandemic, which increases the feasibility of thermal-based crowd counting. However, thermal images are not perfect. As shown in Fig.~\ref{fig:RGBT_sample}-(c,d), some hard negative objects (e.g., heating walls and lamps) are also highlighted in thermal images, but they can be eliminated effectively with the aid of optical information. Overall, RGB images and thermal images are highly complementary.
To the best of our knowledge, no attempts have been made to simultaneously explore RGB and thermal images for estimating the crowd counts. In this work, to promote further researches of this field, we propose a large-scale benchmark ``RGBT Crowd Counting ({\bf{RGBT-CC}})'', which contains 2,030 pairs of RGB-thermal images and 138,389 annotated pedestrians. Moreover, our benchmark makes significant advances in terms of diversity and difficulty, as these RGBT images were captured from unconstrained scenes (e.g., malls, streets, train stations, etc.) with various illumination (e.g., day and night).

Nevertheless, capturing the complementarities of multimodal data (i.e., RGB and thermal images) is non-trivial. Conventional methods \cite{lian2019density,zhou2020cascaded,piao2019depth,jiang2020emph,zhai2020bifurcated,sun2019leveraging} either feed the combination of multimodal data into deep neural networks or directly fuse their features, which could not well exploit the complementary information. In this work, to facilitate the multimodal crowd counting, we introduce a cross-modal collaborative representation learning framework, which incorporates multiple modality-specific branches, a modality-shared branch, and an Information Aggregation-Distribution Module (IADM) to fully capture the complementarities among different modalities.
Specifically, our IADM is integrated with two collaborative components, including {\color{red}\bf{i)}} an Information Aggregation Transfer that dynamically aggregates the contextual information of all modality-specific features to enhance the modality-shared feature and {\color{red}\bf{ii)}} an Information Distribution Transfer that propagates the modality-shared information to symmetrically refine every modality-specific feature for further representation learning. Furthermore, the tailor-designed IADM is embedded in different layers to learn the cross-modal representation hierarchically. Consequently, the proposed framework can generate knowledgeable features with comprehensive information, thereby yielding high-quality crowd density maps.

It is worth noting that our method has three appealing properties. {\bf{First}}, thanks to the dual information propagation mechanism, IADM can effectively capture the multi-modal complementarities to facilitate the crowd counting task. {\bf{Second}}, as a plug-and-play module, IADM can be easily incorporated into various backbone networks for end-to-end optimization. {\bf{Third}}, our framework is universal for multimodal crowd counting. Except for RGBT counting, the proposed method can also be directly applied for RGB-Depth counting. 
In summary, the major contributions of this work are three-fold:
\begin{itemize}
\vspace{-2mm}
\setlength{\itemsep}{0pt}
\setlength{\parsep}{0pt}
\setlength{\parskip}{0pt}
  \item We introduce a large-scale RGBT benchmark to promote the research of crowd counting, in which 138,389 pedestrians are annotated in 2,030 pairs of RGB-thermal images captured in unconstrained scenarios.
  \item We develop a cross-modal collaborative representation learning framework, which is capable of fully learning the complementarities among different modalities with a Information Aggregation-Distribution Module.
  \item Extensive experiments conducted on two multimodal benchmarks (i.e., RGBT-CC and ShanghaiTechRGBD \cite{lian2019density}) greatly demonstrate that the proposed method is effective and universal for multimodal crowd counting.
\end{itemize}

\section{Related Works}

{\bf{Crowd Counting Benchmarks:}}
In recent years, we have witnessed the rapid evolution of crowd counting benchmarks. UCSD \cite{chan2008privacy} and WorldExpo \cite{zhang2015cross} are two early datasets that respectively contain 2,000 and 3,980 video frames with low diversities and low-medium densities. To alleviate the limitations of the aforementioned datasets, Zhang \textit{et al.}~\cite{zhang2016single} collected 1,198 images with 330,165 annotated heads, which are of better diversity in terms of scenes and density levels. Subsequently, three large-scale datasets were proposed in succession. For instance, UCF-QNRF \cite{idrees2018composition} is composed of 1,535 high density images images with a total of 1.25 million pedestrians. JHU-CROWD++ \cite{sindagi2020jhu} contains 4,372 images with 1.51 million annotated heads, while NWPU-Crowd \cite{gao2020nwpu} consists of 2.13 million annotations in 5,109 images. Nevertheless, all the above benchmarks are based on RGB optical images, in which almost all previous methods fail to recognize the invisible pedestrians in poor illumination conditions. Recently, Lian \textit{et al.}~\cite{lian2019density} utilized a stereo camera to capture 2,193 depth images that are insensitive to illumination. However, these images are coarse in outdoor scenes due to the limited depth ranges (0{\small{$\sim$}}20 meters), which seriously restricts their deployment scopes. Fortunately, we find that thermal images are robust to illumination and have large perception distance, thus can help to recognize pedestrians under various scenarios. Therefore, we propose the first RGBT crowd counting dataset in this work, hoping that it would greatly promote the future development in this field.


{\bf{Crowd Counting Approaches:}}
As a classics problem in computer vision, crowd counting has been studied extensively. Early works \cite{Chan2009bayesian,chen2012feature,idrees2013multi} directly predict the crowd count with regression models, while subsequent methods usually generate crowd density maps and then accumulate all pixels' values to obtain the final counts.
Specifically, a large number of deep neural networks with various architectures \cite{fu2015fast,zhang2015cross,wang2015deep,walach2016learning,sam2017switching,kang2017incorporating,sindagi2017generating,li2018csrnet,zhang2019relational,liu2019adcrowdnet,qiu2019crowd,
jiang2019crowd,yuan2020crowd} and loss functions \cite{cao2018scale,idrees2018composition,ma2019bayesian,liu2019crowd} are developed for still image-based crowd counting. Meanwhile, some methods \cite{zhang2019wide,xiong2017spatiotemporal,ren2018fusing,liu2020estimating} perform crowd estimation from multi-view images or surveillance videos.
%
%
%
However, all aforementioned methods estimate crowd counts with the optical information of RGB images/videos and are not effective when working in poor illumination conditions. Recently, depth images are used as auxiliary information to count and locate human heads~\cite{lian2019density}. Nevertheless, depth images are coarse in outdoor scenarios, thus depth-based methods have relatively limited deployment scopes.  Nevertheless, depth images are coarse in outdoor scenarios, thus depth-based methods have relatively limited deployment scopes.

{\bf{Multi-Modal Representation Learning:}}
Multi-modal representation learning aims at comprehending and representing cross-modal data through machine learning.
There are many strategies in cross-modal feature fusion. Some simple fusion methods~\cite{kiela2014learning,lian2019density,sun2019leveraging,fu2020jl} obtain a fused feature with the operations of element-wise multiplication/addition or concatenation in the ``Early Fusion'' and ``Late Fusion'' way. To exploit the advantages of both early and late fusion, various two-stream-based models~\cite{wu2020deepdualmapper,piao2020a2dele,zhao2019contrast,zhang2019attend} propose to fuse hierarchical cross-modal features, achieving the fully representative shared feature. Besides, a few approaches~\cite{lu2020cross} explore the use of a shared branch, mapping the shared information to common feature spaces.
%
Furthermore, some recent works~\cite{fan2020bbsnet,HDFNet-ECCV2020,zhang2020uc} are presented to address RGBD saliency detection, which is also a cross-modal dense prediction task like RGBT crowd counting.
However, most of these works are one-way information transfer, just using depth modality as auxiliary information to help the representation learning of RGB modality. In this work, we propose a symmetric dynamic enhancement mechanism that can take full advantage of the modal complementarities in crowd counting.

\begin{figure}[t]
\centering
    \includegraphics[width=0.375\textwidth]{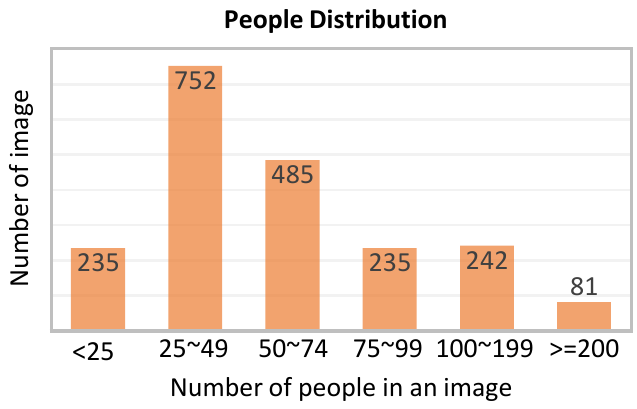}
    \vspace{-1mm}
    \caption{The statistics histogram of people distribution in the proposed RGBT Crowd Counting benchmark.}
    \vspace{0mm}
    \label{fig:distribution}
\end{figure}

\begin{figure*}[t]
\centering
   \includegraphics[width=1.9\columnwidth]{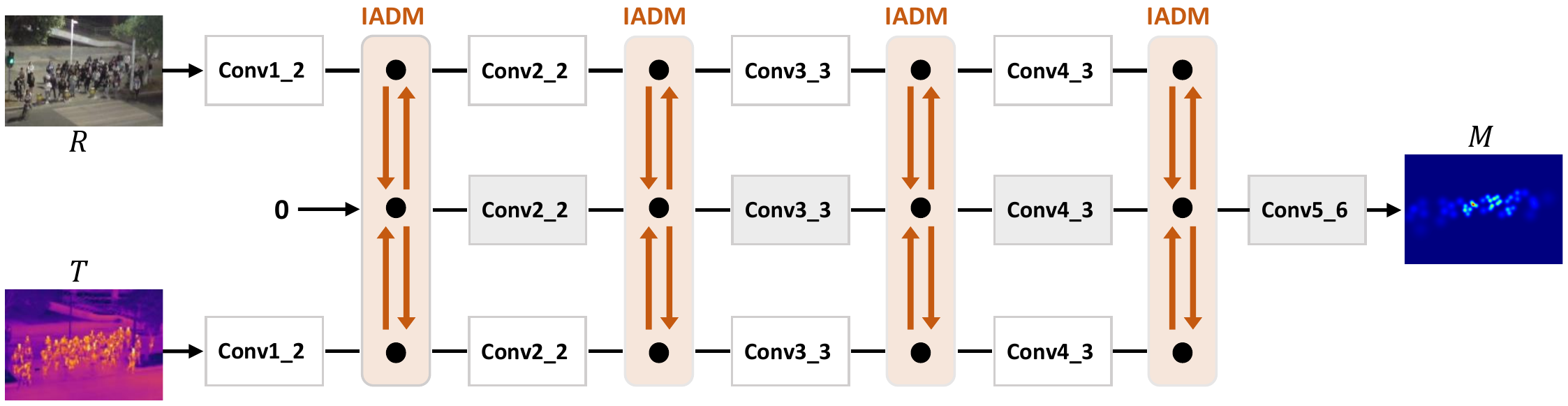}
\vspace{-1mm}
   \caption{The architecture of the proposed cross-modal collaborative representation learning framework for multimodal crowd counting. Specifically, our framework consists of multiple modality-specific branches, a modality-shared branch, and an Information Aggregation-Distribution Module (IADM).
   }
\vspace{-3mm}
\label{fig:framework}
\end{figure*}

\begin{table}
  \caption{The training, validation and testing sets of our RGBT-CC benchmark. In each grid, the first value is the number of images, while the second value denotes the average count per image.}
  \vspace{0mm}
  \centering
    \begin{tabular}{c|c|c|c}
    \hline
      & {{Training}} & {{Validation}} & {{Testing}} \\
    \hline
    \hline
    \#Bright        & {~~}510  / 65.66 & {~~}97 / 63.02 & 406 / 73.39  \\
    \#Dark{~~~}     & {~~}520  / 62.52 & 103 / 67.74 & 394 / 74.88 \\
    \#Total{~~}     & 1030 / 64.07 & 200 / 65.45 & 800 / 74.12 \\
    \hline
    Scene & \multicolumn{3}{c}{malls, streets, train/metro stations, etc}  \\
    \hline
    \end{tabular}
  \label{tab:data_split}
  \vspace{-1mm}
\end{table}

\section{RGBT Crowd Counting Benchmark}
To the best of our knowledge, there is currently no public RGBT dataset for crowd counting. To promote the future research of this task, we propose a large-scale RGBT Crowd Counting ({\bf{RGBT-CC}}) benchmark. Specifically, we first use an optical-thermal camera to take a large number of RGB-thermal images in various scenarios (e.g., malls, streets, playgrounds, train stations, metro stations, etc). Due to the different types of electronic sensors, original RGB images have a high resolution of 2,048$\times$1,536 with a wider field of view, while thermal images have a standard resolution of 640$\times$480 with a smaller field of view. On the basis of coordinate mapping relation, we crop the corresponding RGB regions and resize them to 640$\times$480. We then choose 2,030 pairs of representative RGB-thermal images for manual annotations. Among these samples, 1,013 pairs are captured in the light and 1,017 pairs are in the darkness. A total of 138,389 pedestrians are marked with point annotations, on average 68 people per image. The detailed distribution of people is shown in Fig.~\ref{fig:distribution}. Finally, the proposed RGBT-CC benchmark is randomly divided into three parts. As shown in Table~\ref{tab:data_split}, 1030 pairs are used for training, 200 pairs are for validation and 800 pairs are for testing. Compared with those Internet-based datasets \cite{idrees2018composition,gao2020nwpu,sindagi2020jhu} with serious bias, our RGBT-CC dataset has closer crowd density distribution to realistic cities, since our images are captured in urban scenes with various densities. Therefore, our dataset has wider applications for urban crowd analysis.


\section{Method}
In this work, we propose a cross-modal collaborative representation learning framework for multimodal crowd counting. Specifically, multiple modality-specific branches, a modality-shared branch, and an Information Aggregation-Distribution Module are incorporated to fully capture the complementarities among different modalities with a dual information propagation paradigm.
In this section, we adopt the representative CSRNet~\cite{li2018csrnet} as a backbone network to develop our framework for RGBT crowd counting. It is worth noting that our framework can be implemented with various backbone networks (e.g., MCNN \cite{zhang2016single}, SANet \cite{cao2018scale}, and BL \cite{ma2019bayesian}), and is also universal for multimodal crowd counting, as verified in Section~\ref{RGBD-counting-exp} by directly applying it to the ShanghaiTechRGBD~\cite{lian2019density} dataset.

\subsection{Overview}
As shown in Fig.~\ref{fig:framework}, the proposed RGBT crowd counting framework is composed of three parallel backbones and an Information Aggregation-Distribution Module (IADM). Specifically, the top and bottom backbones are developed for modality-specific (i.e. RGB images and thermal images) representation learning, while the middle backbone is designed for modality-shared representation learning. To fully exploit the multimodal complementarities, our IADM dynamically transfers the specific-shared information to collaboratively enhance the modality-specific and modality-shared representations. Consequently, the final modality-shared feature contains comprehensive information and facilitates generating high-quality crowd density maps.

Given an RGB image $R$ and a thermal image $T$, we first feed them into different branches to extract modality-specific features, which maintain the specific information of individual modality. The modality-shared branch takes a zero-tensor as input\footnote{When the input of modality-shared branch is set to 0, Eq.\ref{eq:aggregation} at Conv$1$\_$2$ is simplified as $\hat{F}_s^{1,2} = I_{r}^{1,2}{\odot}{Conv_{1*1}}(I_{r}^{1,2}) + I_{t}^{1,2}{\odot}{Conv_{1*1}}(I_{t}^{1,2})$. In other words, the initial modality-shared feature is generated by directly aggregating the information of RGB and thermal features.} and aggregates the information of modality-specific features hierarchically. As mentioned above, each branch is implemented with CSRNet, which consists of (1) a front-end block with the first ten convolutional layers of VGG16 \cite{simonyan2014very} and (2) a back-end block with six dilated convolutional layers. More specifically, the modality-specific branches are based on the CSRNet front-end block, while the modality-shared branch is based on the last 14 convolutional layers of CSRNet. In our work, the $j$-th dilated convolutional layer of back-end block is renamed as ``Conv$5$\_$j$''. For convenience, the RGB, thermal, and modality-shared features at Conv$i$\_$j$ layer are denoted as $F_r^{i,j}$, $F_t^{i,j}$, and $F_s^{i,j}$, respectively.

\begin{figure*}
\centering
   \includegraphics[width=1.70\columnwidth]{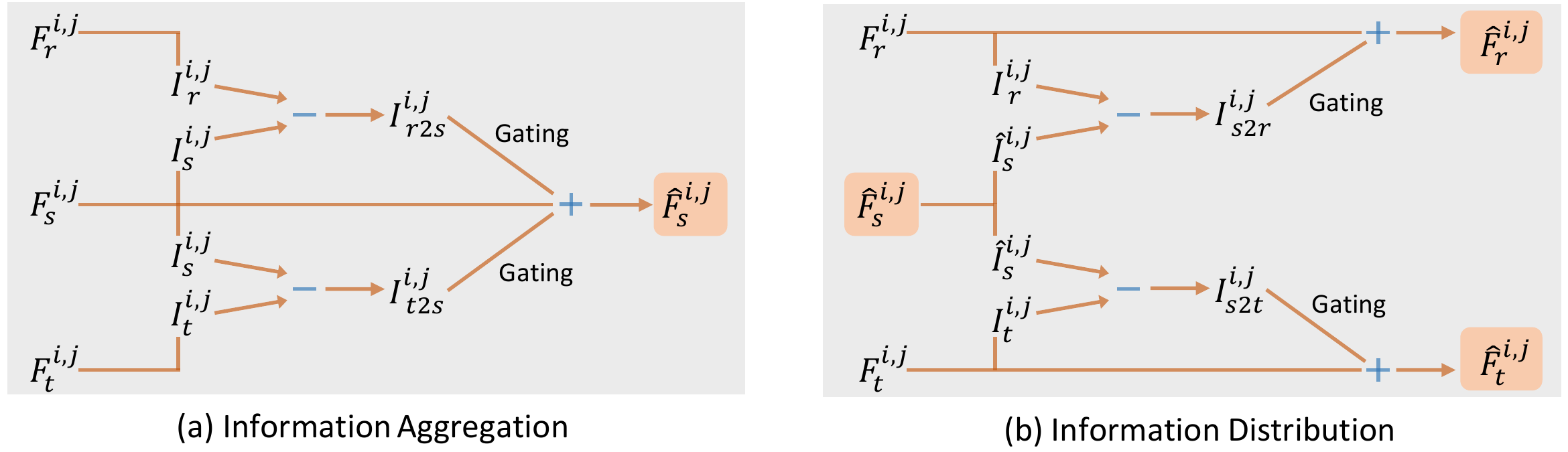}
\vspace{-1mm}
   \caption{{\bf{(a) Information Aggregation Transfer:}} we first extract the contextual information $I_r$/$I_t$ from modality-specific features $F_r$/$F_t$, and then propagate them dynamically to enhance the modality-shared feature ${F}_s$.
   {\bf{(b) Information Distribution Transfer:}} the contextual information $\hat{I}_s$ of the enhance feature $\hat{F}_s$ is distributed adaptively to each modality-specific feature for feedback refinement.
   ``+'' denotes element-wise addition and ``-'' refers to element-wise subtraction.
   }
\vspace{-2mm}
\label{fig:IAD}
\end{figure*}

After feature extraction, we employ the Information Aggregation-Distribution Module described in Section \ref{IADM} to learn cross-modal collaborative representation. To exploit the multimodal information hierarchically, the proposed IADM is embedded after different layers, such as Conv$1$\_$2$, Conv$2$\_$2$, Conv$3$\_$3$, and Conv$4$\_$3$. Specifically, after Conv$i$\_$j$, IADM dynamically transfers complementary information among modality-specific and modality-shared features for mutual enhancement. This process can be formulated as follow:
{
\setlength\abovedisplayskip{5pt}
\setlength\belowdisplayskip{5pt}
\begin{equation}
\hat{F}_s^{i,j}, \hat{F}_r^{i,j}, \hat{F}_t^{i,j} = \text{IADM}(F_s^{i,j}, F_r^{i,j}, F_t^{i,j}), 
\end{equation}
}%
where $\hat{F}_s^{i,j}$, $\hat{F}_r^{i,j}$, and $\hat{F}_t^{i,j}$ are the enhanced features of ${F}_s^{i,j}$, ${F}_r^{i,j}$, and ${F}_t^{i,j}$ respectively. These features are then fed into the next layer of each branch to further learn high-level multimodal  representations. Thanks to the tailor-designed IADM, the complementary information of the input RGB image and the thermal image is progressively transferred into the modality-shared representations. The final modality-shared feature ${F}_s^{5,6}$ contains rich information. Finally, we directly feed ${F}_s^{5,6}$  into a 1*1 convolutional layer for prediction of the crowd density map $M$.

\subsection{Collaborative Representation Learning}\label{IADM}
As analyzed in Section \ref{intro}, RGB images and thermal images are highly complementary. To fully capture their complementarities, we propose an Information Aggregation and Distribution Module (IADM) to collaboratively learn cross-modal representation with a dual information propagation mechanism. Specifically, our IADM is integrated with two collaborative transfers, which dynamically propagate the contextual information to mutually enhance the modality-specific and modality-shared representations.

{\bf{1) Contextual Information Extraction: }}
In this module, we propagate the contextual information rather than the original features, because the later manner would cause the excessive mixing of specific-shared features. To this end, we employ a $L$-level pyramid pooling layer to extract the contextual information for a given feature $F^{i,j} \in R^{h{\times}w{\times}c}$.
Specifically, at the $l^{th}$ level ($l$=1,...,$L$), we apply a $2^{l-1}{\times}2^{l-1}$ max-pooling layer to generate a $\frac{h}{2^{l-1}}{\times}\frac{w}{2^{l-1}}$ feature, which is then upsampled to $h{\times}w$ with nearest neighbor interpolation. For convenience, the upsampled feature is denoted as $F^{i,j,l}$. Finally, the contextual information $I^{i,j} \in R^{h{\times}w{\times}c}$ of feature $F^{i,j}$ is computed as:
{
\setlength\abovedisplayskip{4pt}
\setlength\belowdisplayskip{4pt}
\begin{equation}
I^{i,j} = {Conv}_{1*1}(F^{i,j,1} \oplus F^{i,j,2} \oplus ... \oplus F^{i,j,L}),
\label{eq:infromation}
\end{equation}
}%
where $\oplus$ denotes an operation of feature concatenation and ${Conv}_{1*1}$ is a 1*1 convolutional layer. This extraction has two advantages. First, with a larger receptive field, each position at $I^{i,j}$ contains more context. Second, captured by different sensors, RGB images and thermal images are not strictly aligned, as shown in Figure \ref{fig:RGBT_sample}. Thanks to the translation invariance of max-pooling layers, we can eliminate the misalignment of RGB-thermal images to some extent.

{\bf{2) Information Aggregation Transfer (IAT): }}
In our work, IAT is proposed to aggregate the contextual information of all modality-specific features to enhance the modality-shared feature. As shown in Fig.~\ref{fig:IAD}-(a), instead of directly absorbing all information, our IAT transfers the complementary information dynamically with a gating mechanism that adaptively filters useful information.
Specifically, given features $F_r^{i,j}$, $F_t^{i,j}$ and $F_s^{i,j}$, we first extract their contextual information $I_r^{i,j}$, $I_t^{i,j}$, and $I_s^{i,j}$ with Eq.~\ref{eq:infromation}. Similar to \cite{zhang2019residual,zhao2019spatiotemporal}, we then obtain two residual information $I_{r2s}^{i,j}$ and $I_{t2s}^{i,j}$ by computing the differences between $I_r^{i,j}$/$I_t^{i,j}$ and $I_s^{i,j}$. Finally, we apply two gating functions to adaptively propagate the complementary information for refining the modality-shared feature ${F}_s^{i,j}$. The enhanced feature $\hat{F}_s^{i,j}$ is formulated as follow:
{
\setlength\abovedisplayskip{4pt}
\setlength\belowdisplayskip{4pt}
\begin{equation}
\begin{split}
I_{r2s}^{i,j} & = I_r^{i,j} - I_s^{i,j}, {~~} w_{r2s}^{i,j} = {Conv_{1*1}}(I_{r2s}^{i,j}), \\
I_{t2s}^{i,j} & = I_t^{i,j} - I_s^{i,j}, {~~} w_{t2s}^{i,j} = {Conv_{1*1}}(I_{t2s}^{i,j}), \\
\hat{F}_s^{i,j} &= F_s^{i,j} + I_{r2s}^{i,j}{\odot}w_{r2s}^{i,j} + I_{t2s}^{i,j}{\odot}w_{t2s}^{i,j},
\end{split}
\label{eq:aggregation}
\end{equation}
}%
where the gating functions are implemented by convolutional layers, $w_{r2s}^{i,j}$ and $w_{t2s}^{i,j}$ are the gating weights. $\odot$ refers to an operation of element-wise multiplication. With such a mechanism, the complementary information is effectively embedded into the modality-shared representation, thus our method can better exploit the multimodal data.

\begin{table*}[t]
  \centering
  \caption{The performance of different inputs and different representation learning approaches on our RGBT-CC benchmark.}
  \label{tab:ab_study}
  \vspace{0.5mm}
  \resizebox{0.675\textheight}{!}{%
  \begin{tabular}{c|c|c|c|c|c|c}
    \hline
    Input Data & Representation Learning & GAME(0) $\downarrow$ & GAME(1) $\downarrow$ & GAME(2) $\downarrow$ & GAME(3) $\downarrow$ & RMSE $\downarrow$  \\
    \hline
    \hline
    RGB                              & - & 33.94 & 40.76 & 47.31 & 57.20 & 69.59\\
    \hline
    T                                & - & 21.64 & 26.22 & 31.65 & 38.66 & 37.38 \\
    \hline
    \multirow{6}{*}{RGBT} & Early Fusion & 20.40 & 23.58 & 28.03 & 35.51 & 35.26 \\
                          & Late fusion  & 19.87 & 25.60 & 31.93 & 41.60 & 35.09 \\
    \cline{2-7}
     & W/O Gating Mechanism              & 19.76 & 23.60 & 28.66 & 36.21 & 33.61 \\
     & W/O Modality-Shared Feature       & 18.67 & 22.67 & 27.95 & 36.04 & 33.73 \\
     & W/O Information Distribution      & 18.59 & 23.08 & 28.73 & 36.74 & 32.91 \\
     & IADM                              & {\bf\color{red}17.94} & {\bf\color{red}21.44} & {\bf\color{red}26.17} & {\bf\color{red}33.33} & {\bf\color{red}30.91} \\
    \hline
  \end{tabular}
  }
\vspace{-1mm}
\end{table*}%

\begin{table*}[t]
  \centering
  \caption{The performance under different illumination conditions on our RGBT-CC benchmark. The unimodal data is directly fed into CSRNet, while the multimodal data is fed into our proposed framework based on CSRNet.}
  \vspace{0.5mm}
  \label{tab:illumination}
  \resizebox{0.60\textheight}{!}{%
  \begin{tabular}{c|c|c|c|c|c|c}
    \hline
    Illumination & Input Data & GAME(0) $\downarrow$ & GAME(1) $\downarrow$ & GAME(2) $\downarrow$ & GAME(3) $\downarrow$ & RMSE $\downarrow$ \\
    \hline
    \hline
    \multirow{3}{*}{Brightness} & RGB  & 23.49 & 30.14 & 37.47 & 48.46 & 45.40\\
                                &  T   & 25.21 & 28.98 & 34.82 & 42.25 & 40.60 \\
                                & RGBT & {\bf\color{red}20.36} & {\bf\color{red}23.57} & {\bf\color{red}28.49}
                                       & {\bf\color{red}36.29} & {\bf\color{red}32.57} \\
    \hline
    \multirow{3}{*}{Darkness}   & RGB  & 44.72 & 51.70 & 57.45 & 66.21 & 87.81 \\
                                &  T   & 17.97 & 23.38 & 28.39 & 34.95 & 33.74 \\
                                & RGBT & {\bf\color{red}15.44} & {\bf\color{red}19.23} & {\bf\color{red}23.79}
                                       & {\bf\color{red}30.28} & {\bf\color{red}29.11} \\
    \hline
  \end{tabular}
  }
  \vspace{-2mm}
\end{table*}%

{\bf{3) Information Distribution Transfer (IDT): }}
After information aggregation, we distribute the information of the new modality-shared feature to refine each modality-specific feature respectively. As shown in Fig.~\ref{fig:IAD}-(b), with the enhanced feature $\hat{F}_s^{i,j}$, we first extract its contextual information $\hat{I}_s^{i,j}$, which is then dynamically propagated to $F_{r}^{i,j}$ and $F_{t}^{i,j}$. Simialr to IAT, two gating functions are used for information filtering.
Specifically, the enhanced modality-specific features are computed as follow:
{
\setlength\abovedisplayskip{4pt}
\setlength\belowdisplayskip{4pt}
\begin{equation}\nonumber
\begin{split}
I_{s2r}^{i,j} &= \hat{I}_s^{i,j} - I_r^{i,j}, {~~~~~~~~~~~~~~~~~} I_{s2t}^{i,j} = \hat{I}_s^{i,j} - I_t^{i,j}, \\
w_{s2r}^{i,j} &= {Conv_{1*1}}(I_{s2r}^{i,j}), {~~~~~~~~} w_{s2t}^{i,j} = {Conv_{1*1}}(I_{s2t}^{i,j}), \\
\hat{F}_r^{i,j} &= F_r^{i,j} + I_{s2r}^{i,j}{\odot}w_{s2r}^{i,j}, {~~~}\hat{F}_t^{i,j} = F_t^{i,j} + I_{s2t}^{i,j}{\odot}w_{s2t}^{i,j}.
\end{split}
\end{equation}
}%
Finally, all enhanced features $\hat{F}_r^{i,j}$, $\hat{F}_t^{i,j}$, and $\hat{F}_s^{i,j}$ are fed into the following layers of the individual branch for further representation learning.

\section{Experiments}\label{experiment}

\subsection{Implementation Details \& Evaluation Metrics}
In this work, the proposed method is implemented with PyTorch~\cite{paszke2019pytorch}. Here we take various models (e.g., CSRNet~\cite{li2018csrnet}, MCNN \cite{zhang2016single}, SANet \cite{cao2018scale}, and BL \cite{ma2019bayesian}) as backbone to develop multiple instances of our framework. To maintain a similar number of parameters to original models for fair comparisons, the channel number of these backbones in our framework is respectively set to 70\%, 60\%, 60\%, and 60\% of their original values. The kernel parameters are initialized by Gaussian distribution with a zero mean and a standard deviation of 1e-2. At each iteration, a pair of 640$\times$480 RGBT image is fed into the network. The ground-truth density map is generated with geometry-adaptive Gaussian kernels \cite{zhang2016single}. The learning rate is set to 1e-5 and Adam \cite{kingma2014adam} is used to optimize our framework. Notice that the loss function of our framework is the same as that of the adopted backbone network.

Following~\cite{liu2020weighing,sindagi2019multi,liu2020efficient}, we adopt the Root Mean Square Error (RMSE) as an evaluation metric. Moreover, Grid Average Mean Absolute Error (GAME \cite{guerrero2015extremely}) is utilized to evaluate the performance in different regions. Specifically, for a specific level $l$, we divide the given images into $4^l$ non-overlapping regions and measure the counting error in each region. Finally, the GAME at level $l$ is computed as:
{
\setlength\abovedisplayskip{0.25mm}
\setlength\belowdisplayskip{0.25mm}
\begin{equation}
GAME(l) = \frac{1}{N} \sum_{i=1}^{N} \sum_{j=1}^{4^l} |\hat{P}_i^j - P_{i}^j|,
\end{equation}
}%
where $N$ is the total number of the testing samples, $\hat{P}_i^j$ and $P_{i}^j$ are the estimated count and the corresponding ground-truth count in the $j^{th}$ region of the $i^{th}$ image. Note that GAME(0) is equivalent to Mean Absolute Error (MAE).

\subsection{Ablation Studies}
We perform extensive ablation studies to verify the effectiveness of each component in our framework. In this subsection, CSRNet is utilized as the backbone network to implement our proposed method.

\begin{figure*}
\centering
   \includegraphics[width=2.1\columnwidth]{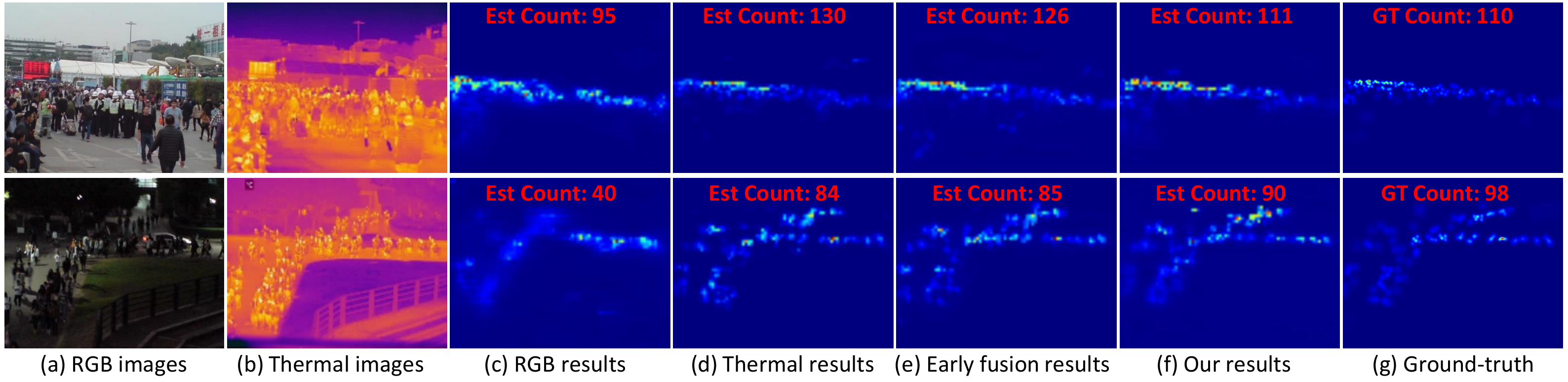}
\vspace{-5mm}
   \caption{Visualization of the crowd density maps generated in different illumination conditions. (a) and (b) show the input RGB images and thermal images. (c) and (d) are the results of RGB-based CSRNet and thermal-based CSRNet. (e) shows the results of CSRNet that takes the concatenation of RGB and thermal images as input. (f) refers to the results of our CSRNet+IDAM. And the ground-truths are shown in (g). We can observe that our density maps and estimated counts are more accurate than those of other methods. (\textit{Best to zoom in to view this figure.})
   }
\label{fig:RGBT_results}
\vspace{0mm}
\end{figure*}

\begin{table*}[t]
  \centering
  \caption{Performance of different methods on the proposed RGBT-CC benchmark. All the methods in this table utilize both RGB images and thermal images to estimate the crowd counts.}
  \resizebox{0.525\textheight}{!}{%
    \begin{tabular}{c|c|c|c|c|c}
    \hline
    Backbone & GAME(0) $\downarrow$ & GAME(1) $\downarrow$ & GAME(2) $\downarrow$  & GAME(3) $\downarrow$ & RMSE $\downarrow$\\
    \hline\hline
    UCNet  \cite{zhang2020uc}     & 33.96 & 42.42 & 53.06 & 65.07 & 56.31 \\
    HDFNet \cite{HDFNet-ECCV2020} & 22.36 & 27.79 & 33.68 & 42.48 & 33.93 \\
    BBSNet \cite{fan2020bbsnet}   & 19.56 & 25.07 & 31.25 & 39.24 & 32.48 \\
    MVMS \cite{zhang2019wide}   & 19.97 & 25.10 & 31.02 & 38.91 & 33.97 \\
    \hline

    {~~}MCNN{~~~~~~~~~~~~~~~} & 21.89 & 25.70 & 30.22 & 37.19 & 37.44 \\
    {~~}MCNN + IADM & {\bf\color{red}19.77} & {\bf\color{red}23.80} & {\bf\color{red}28.58} & {\bf\color{red}35.11} & {\bf\color{red}30.34} \\
    \hline

    {~~}SANet{~~~~~~~~~~~~~~~} & 21.99 & 24.76 & 28.52 & 34.25 & 41.60 \\
    {~~}SANet + IADM & {\bf\color{red}18.18} & {\bf\color{red}21.84} & {\bf\color{red}26.27} & {\bf\color{red}32.95} & {\bf\color{red}33.72}\\
    \hline

    CSRNet{~~~~~~~~~~~~~~~} & 20.40 & 23.58 & 28.03 & 35.51 & 35.26 \\
    CSRNet + IADM & {\bf\color{red}17.94} & {\bf\color{red}21.44} & {\bf\color{red}26.17} & {\bf\color{red}33.33} & {\bf\color{red}30.91} \\
    \hline

    {~~~~~~~~}BL{~~~~~~~~~~~~~~~} & 18.70 & 22.55 & 26.83 & 34.62 & 32.67 \\
    {~~~~~~~~}BL + IADM & {\bf\color{red}15.61} & {\bf\color{red}19.95} & {\bf\color{red}24.69} & {\bf\color{red}32.89} & {\bf\color{red}28.18} \\
    \hline
    \end{tabular}
  }
    \vspace{-2mm}
    \label{tab:RGBT_result}
\end{table*}

{\bf{1) Effectiveness of Multimodal Data:}}
We first explore whether the multimodal data (i.e., RGB images and thermal images) is effective for crowd counting. As shown in Table~\ref{tab:ab_study}, when only feeding RGB images into CSRNet, we obtain less impressive performance (e.g., GAME(0) is 33.94 and RMSE is 69.59), because we cannot effectively recognize people in dark environments. When utilizing thermal images, GAME(0) and RMSE are sharply reduced to 21.64 and 37.38, which demonstrates that thermal images are more useful than RGB images. In contrast, various models in the bottom six rows of Table \ref{tab:ab_study} achieve better performance, when considering RGB and thermal images simultaneously. In particular, our CSRNet+IADM has a relative performance improvement of 17.3\% on RMSE, compared with the thermal-based CSRNet.

To further verify the complementarities of multimodal data, the testing set is divided into two parts to measure the performance in different illumination conditions separately. As shown in Table~\ref{tab:illumination}, using both RGB and thermal images, our CSRNet+IADM consistently outperforms the unimodal CSRNet in both bright and dark scenarios. This is attributed to the thermal information that greatly helps to distinguish potential pedestrians from the cluttered background, while optical information is beneficial to eliminate heating negative objects in thermal images. Moreover, we also visualize some crowd density maps generated with different modal data in Fig.~\ref{tab:RGBT_result}. We can observe that the density maps and estimated counts of our CSRNet+IADM are more accurate. These quantitative and qualitative experiments show that RGBT images are greatly effective for crowd counting.

\begin{table}[t]
  \centering
\caption{Performance of different level numbers of the pyramid pooling layer in IADM.}
  \label{tab:level_number}
  \resizebox{8.3cm}{!}{%
  \begin{tabular}{c|c|c|c|c|c}
    \hline
    \#Level & GAME(0) & GAME(1) & GAME(2) & GAME(3) & RMSE \\
    \hline\hline
    $L$=1 & 18.94 & 23.05 & 28.03 & 35.88 & 33.01 \\
    $L$=2 & 18.35 & 22.56 & 27.84 & 35.90 & 31.94 \\
    $L$=3 & {\bf\color{red}17.94} & {\bf\color{red}21.44} & {\bf\color{red}26.17} & {\bf\color{red}33.33} & {\bf\color{red}30.91} \\
    $L$=4 & 17.80 & 21.39 & 25.91 & 33.20 & 31.48 \\
    \hline
  \end{tabular}
  }
  \vspace{-4mm}
\end{table}%

\begin{table*}[t]
  \centering
      \caption{Performance of different methods on the ShanghaiTechRGBD benchmark. All the methods in this table utilize both RGB images and depth images to estimate the crowd counts.}
  \resizebox{0.525\textheight}{!}{%
    \begin{tabular}{c|c|c|c|c|c}
    \hline
    Method & GAME(0) $\downarrow$ & GAME(1) $\downarrow$ & GAME(2) $\downarrow$  & GAME(3) $\downarrow$ & RMSE $\downarrow$\\
    \hline\hline
    UCNet  \cite{zhang2020uc}     & 10.81 & 15.24 & 22.04 & 32.98 & 15.70 \\
    HDFNet \cite{HDFNet-ECCV2020} & 8.32  & 13.93 & 17.97 & 22.62 & 13.01 \\
    BBSNet \cite{fan2020bbsnet}   & 6.26  & 8.53 & 11.80 & 16.46 & 9.26 \\
    \hline

    DetNet  \cite{liu2018decidenet} & 9.74 & - & - & - & 13.14 \\
    CL \cite{idrees2018composition} & 7.32 & - & - & - & 10.48 \\
    RDNet \cite{lian2019density}    & 4.96 & - & - & - & 7.22 \\
    \hline

    {~~}MCNN{~~~~~~~~~~~~~~~} & 11.12 & 14.53 & 18.68 & 24.49 & 16.49 \\
    {~~}MCNN + IADM         & {\bf\color{red}9.61}  & {\bf\color{red}11.89} & {\bf\color{red}15.44} & {\bf\color{red}20.69} & {\bf\color{red}14.52} \\
    \hline

    {~~~~~~~~}BL{~~~~~~~~~~~~~~~} & 8.94  & 11.57 & 15.68 & 22.49 & 12.49 \\
    {~~~~~~~~}BL + IADM  & {\bf\color{red}7.13}  & {\bf\color{red}9.28}  & {\bf\color{red}13.00} & {\bf\color{red}19.53} & {\bf\color{red}10.27} \\
    \hline

    {~~}SANet{~~~~~~~~~~~~~~~}  & 5.74  & 7.84  & 10.47 & 14.30 & 8.66 \\
    {~~}SANet + IADM  & {\bf\color{red}4.71}  & {\bf\color{red}6.49}  & {\bf\color{red}9.02}  & {\bf\color{red}12.41}	& {\bf\color{red}7.35}\\
    \hline

    CSRNet{~~~~~~~~~~~~~~~} & 4.92  & 6.78  & 9.47  & 13.06 & 7.41 \\
    CSRNet + IADM & {\bf\color{red}4.38}  & {\bf\color{red}5.95}  & {\bf\color{red}8.02}  & {\bf\color{red}11.02} & {\bf\color{red}7.06} \\
    \hline
    \end{tabular}
  }
    \vspace{-3mm}
    \label{tab:RGBD_result}
\end{table*}

{\bf{2) Which Representation Learning Method is Better?}}
We implement six methods for multimodal representation learning. Specifically, ``Early Fusion'' feeds the concatenation of RGB and thermal images into CSRNet. ``Late Fusion'' extracts the RGB and thermal features respectively with two CSRNet and then combines their features to generate density maps. As shown in Table~\ref{tab:ab_study}, these two models are better than unimodal models, but their performance still lags far behind various variants of our IADM. For instance, without gating functions, the variant ``W/O Gating Mechanism'' directly propagates information among different features and obtains an RMSE of 33.61. The variant ``W/O Modality-Shared Feature'' obtains a GAME(0) of 18.67 and an RMSE of 33.73, when removing the modality-shared branch and directly refining the modality-specific features. When using the modality-shared branch but only aggregating multimodal information, the variant ``W/O Information Distribution'' obtains a relatively better RMSE 32.91. When using the full IADM, our method achieves the best performance on all evaluation metrics. This is attributed to our tailor-designed architecture (i.e., specific-shared branches, dual information propagation) that can effectively learn the multimodal collaborative representation, and fully capture the complementary information of RGB and thermal images. These experiments demonstrate the effectiveness of the proposed IADM for multimodal representation learning.

{\bf{3) The Effectiveness of Level Number of Pyramid Pooling Layer: }}
In the proposed IDAM, an $L$-level pyramid pooling layer is utilized to extract contextual information. In this section, we explore the effectiveness of the level number. As shown in Table \ref{tab:level_number}, when $L$ is set to 1, the GAME(3) and RMSE are 35.88 and 33.01, respectively. As the level number increases, our performance also becomes better gradually, and we can achieve very competitive results when the pyramid pooling layer has three levels. More levels over 3 will not bring additional performance gains. Thus, the level number $L$ is consistently set to 3 in our work.

\subsection{Comparison with State-of-the-Art Methods}
We compare the proposed method with state-of-the-art methods on the large-scale RGBT-CC benchmark. The compared methods can be divided into two categories. The first class is the specially-designed models for crowd counting, including MCNN \cite{zhang2016single}, SANet \cite{cao2018scale}, CSRNet~\cite{li2018csrnet}, and BL \cite{ma2019bayesian}. These methods are reimplemented to take the concatenation of RGB and thermal images as input in an ``Early Fusion" way. Moreover, MVMS \cite{zhang2019wide} is also reimplemented on RGBT-CC and pixel-wise attention map \cite{chen2016attention} is utilized to fuse the features of optical view and thermal view. The second class is several best-performing models for multimodal learning, including UCNet \cite{zhang2020uc}, HDFNet \cite{HDFNet-ECCV2020}, and BBSNet \cite{fan2020bbsnet}. Based on their official codes, these methods are reimplemented to estimate crowd counts on our RGBT-CC dataset. As mentioned above, our IADM can be incorporated into various networks, thus here we take CSRNet, MCNN, SANet, and BL as backbone to develop multiple instances of our framework.

The performance of all comparison methods is shown in Table \ref{tab:RGBT_result}. As can be observed, all instances of our method outperform the corresponding backbone networks consistently. For instance, both MCNN+IADM and SANet+IADM have a relative performance improvement of 18.9\% on RMSE, compared with their ``Early Fusion'' models. Moreover, our CSRNet+IADM and BL+IADM achieve better performance on all evaluation metrics, compared with other advanced methods (i.e., UCNet, HDFNet, and BBSNet). This is because our method learns specific-shared representations explicitly and enhances them mutually, while others just simply fuse multimodal features without mutual enhancements. Thus our method can better capture the complementarities of RGB images and thermal images. This comparison has demonstrated the effectiveness of our method for RGBT crowd counting.

\subsection{Apply to RGBD Crowd Counting}\label{RGBD-counting-exp}
We apply the proposed method to estimate crowd counts from RGB images and depth images. In this subsection, we also take various crowd counting models as backbone to develop our framework on ShanghaiTechRGBD \cite{lian2019density} benchmark. The implementation details of the compared methods are similar to the previous subsection. As shown in Table~\ref{tab:RGBD_result}, all instances of our framework are superior to their corresponding backbone networks by obvious margins. Moreover, our SANet+IADM and CSRNet+IADM outperform three advanced models (i.e., UCNet, HDFNet, and BBSNet) on all evaluation metrics. More importantly, our CSRNet+IADM achieves the lowest GAME(0) 4.38 and RMSE 7.05, and becomes the new state-of-the-art method on ShanghaiTechRGBD benchmark. This experiment shows that our approach is universal and effective for RGBD crowd counting.

\section{Conclusion}
In this work, we propose to incorporate optical and thermal information to estimate crowd counts in unconstrained scenarios. To this end, we introduce the first RGBT crowd counting benchmark with 2,030 pairs of RGB-thermal images and 138,389 annotated people. Moreover, we develop a cross-modal collaborative representation learning framework, which utilizes a tailor-designed Information Aggregation-Distribution Module to fully capture the complementary information of different modalities. Extensive experiments on two real-world benchmarks show the effectiveness and universality of the proposed method for multimodal (e.g., RGBT and RGBD) crowd counting.

\section*{Acknowledgments}
This work was supported
in part by National Natural Science Foundation of China under Grant No.U1811463,
in part by National Key R\&D Program of China under Grant No.2020AAA0109700,
in part by National Natural Science Foundation of China under Grant No.61976250 and 61876045,
in part by the Guangdong Basic and Applied Basic Research Foundation under Grant No.2017A030312006 and 2020B1515020048,
in part by Major Project of Guangzhou Science and Technology of Collaborative Innovation and Industry under Grant 201605122151511.


{\small
\bibliographystyle{ieee_fullname}
\bibliography{egbib}
}

\end{document}